\documentclass[letterpaper, 10 pt, conference]{ieeeconf}  

\IEEEoverridecommandlockouts                              

\usepackage{epsfig} 
\usepackage{amsmath} 
\usepackage{amssymb}  
\usepackage{cite}
\usepackage[linesnumbered,ruled]{algorithm2e}
\usepackage[normalem]{ulem} 
\usepackage{hyperref}
\usepackage{booktabs}
\usepackage{multirow}
\usepackage[caption=false,font=footnotesize]{subfig}
\usepackage[dvipsnames]{xcolor}

\usepackage{soul}
\newcounter{RNum}
\renewcommand{\theRNum}{\arabic{RNum}}
\newcommand{\Remark}{\noindent\textit{\textbf{Remark}~\refstepcounter{RNum}\textbf{\theRNum}: }}
\newcommand{\NoOne}[1]{\textcolor{red}{#1}}
\newcommand{\NoTwo}[1]{\textcolor{green}{#1}}
\newcommand{\NoThree}[1]{\textcolor{blue}{#1}}
\soulregister\NoOne7
\soulregister\NoTwo7
\soulregister\NoThree7
\soulregister\Remark7

\title{\LARGE \bf
Ad$^2$Attack: Adaptive Adversarial Attack on Real-Time UAV Tracking
}

\author{Changhong Fu$^{1,*}$, Sihang Li$^{1}$, Xinnan Yuan$^{1}$, Junjie Ye$^{1}$, Ziang Cao$^{2}$, and Fangqiang Ding$^{3}$ 
\thanks{$^{*}$Corresponding author}%
\thanks{$^{1}$Changhong Fu, Sihang Li, Xinnan Yuan, and Junjie Ye are with the School of Mechanical Engineering, Tongji University, Shanghai 201804, China.
        {\tt\footnotesize changhongfu@tongji.edu.cn}}%
\thanks{$^{2}$Ziang Cao is with the School of Automotive Studies, Tongji University, Shanghai 201804, China.}
\thanks{$^{3}$Fangqiang Ding is with the School of Informatics at the University of Edinburgh, United Kingdom EH8 9YL, UK.}
}

\begin{document}

\maketitle
\thispagestyle{empty}
\pagestyle{empty}

\begin{abstract}
Visual tracking is adopted to extensive unmanned aerial vehicle (UAV)-related applications, which leads to a highly demanding requirement on the robustness of UAV trackers. 
However, adding imperceptible perturbations can easily fool the tracker and cause tracking failures. This risk is often overlooked and rarely researched at present. Therefore, to help increase awareness of the potential risk and the robustness of UAV tracking,
this work proposes a novel adaptive adversarial attack approach, \textit{i.e.}, Ad$^2$Attack, against UAV object tracking.
Specifically, adversarial examples are generated online during the resampling of the search patch image, which leads trackers to lose the target in the following frames. 
Ad$^2$Attack is composed of a direct downsampling module and a super-resolution upsampling module with adaptive stages. 
A novel optimization function is proposed for balancing the imperceptibility and efficiency of the attack.
Comprehensive experiments on several well-known benchmarks and real-world conditions show the effectiveness of our attack method, which dramatically reduces the performance of the most advanced Siamese trackers.
\end{abstract}



\section{Introduction} \label{sec:intro}

Unmanned aerial vehicle (UAV) tracking aims to locate the indicated object based on its initial state from an aerial perspective in the subsequent frames. With the powerful flexibility of UAV, UAV tracking in numerous applications\cite{bonatti2019IROS}, \cite{yu2020autonomous}, \cite{lopez2021framework} has garnered considerable attention in recent years. 
Generally, the methods of UAV tracking are categorized as correlation filter (CF)-based trackers \cite{huang2019learning}, \cite{li2020autotrack}, \cite{li2020training} and deep learning (DL)-based trackers \cite{li2020keyfilter}, \cite{bertinetto2016fully}. Although the former is energy-saving and less expensive in computation, the performance remains significantly inferior to that of the latter.
In recent years, approaches based on DL have made substantial progress in UAV tracking\cite{Cao2021IROS}, \cite{fu2021onboard}, \cite{cao2021hift}. The most eye-catching work is the trackers based on the Siamese network. 
Due to their increased speed and precision, these trackers are extensively employed on UAV platforms.
However, due to the susceptibility of deep neural networks (DNNs) to adversarial examples \cite{goodfellow2014explaining}, Siamese trackers are easy to be attacked by the small perturbations on the input image, resulting in locating a wrong position, which is a huge hazard to the tasks based on UAV tracking and is worthy of attention.

The adversarial attack aims to make the DNN model output the wrong prediction by adding small perturbations that are imperceptible to human eyes to the images\cite{szegedy2014intriguing}. Specific to the field of visual tracking,
the most widely-used adversarial attack approaches can be mainly classified into two categories: \textit{1)} attacks based on iterative optimization and \textit{2)} attacks based on DNNs.
The former method\cite{wiyatno2019physical} aims to use perceptible texture to mislead the GOTURN tracker\cite{held2016learning}. However, this method hardly achieves real-time attack due to the heavy computation during iterative optimization. Albeit some attack method\cite{wiyatno2019physical} is efficient, it needs the attacker to change the surroundings physically in real-world.
The latter methods\cite{yan2020cooling},\cite{chen2020one} offline train adversarial perturbation generators using massive amounts of data. 
To maximize the attack's effectiveness, this category inputs the original resolution image into the generator, which undoubtedly increases the amount of calculation, and also makes the perturbation in the whole image easy to be noticed.
Therefore, the existing two categories are not suitable for the task of attacking UAV tracking, which requires real-time high performance.

This work proposes a real-time adversarial attack method, \textit{i.e.}, Ad$^2$Attack, against the state-of-the-art (SOTA) trackers based on the Siamese network\cite{zhang2019deeper}.
Inspired by image-resample technology\cite{sun2020learned}, \cite{xiao2020invertible}, rather crafting adversarial examples by adding adversarial perturbation on original images, this method adaptively learns a complicated adversarial mapping from low-resolution (LR) image to original high-resolution (HR) image, with the goal of resampling the image to blind the tracker and minimizing computational costs.
\begin{figure}[!t]	
	\centering
	\includegraphics[width=1\linewidth]{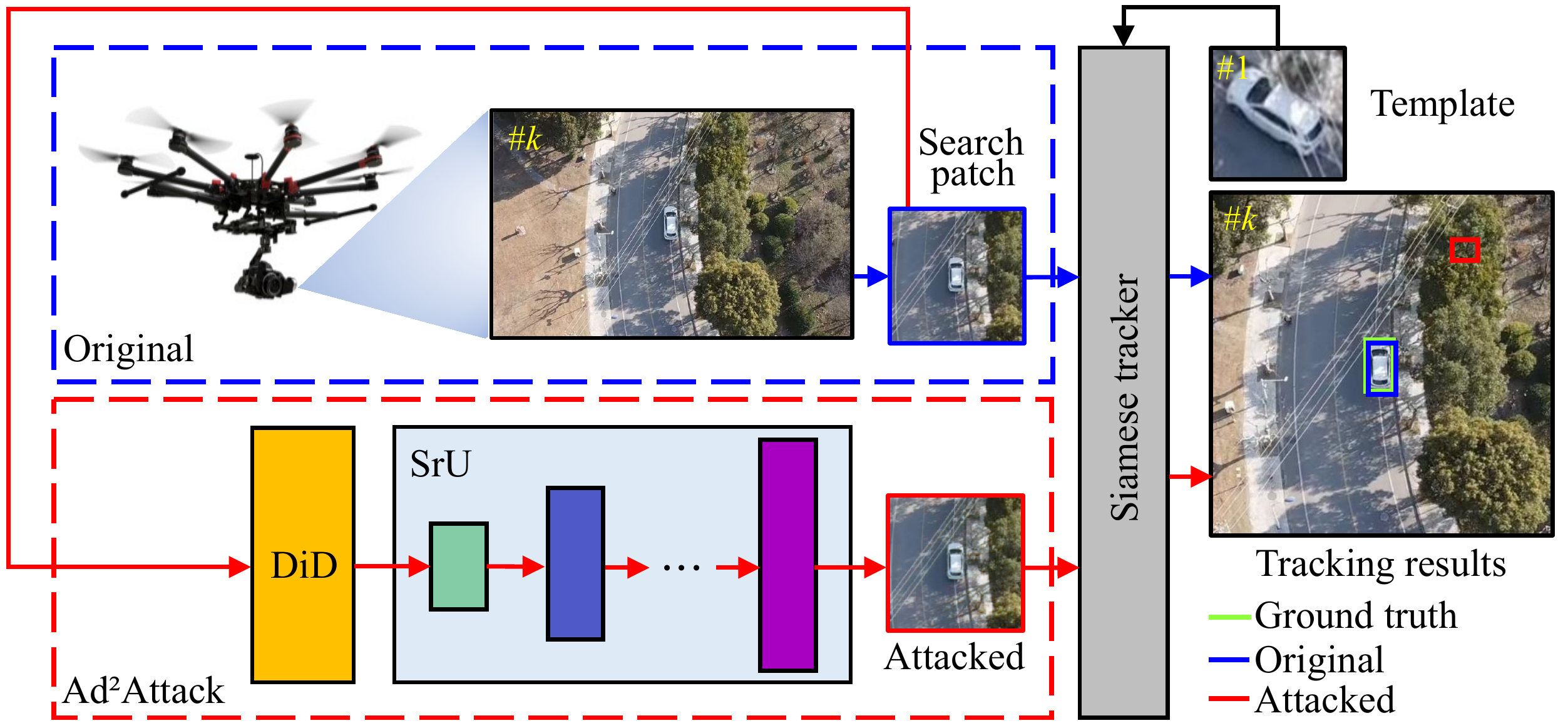}
	\setlength{\abovecaptionskip}{-14pt} 
	\caption
	{   Overall comparison of tracking results from a Siamese            tracker, with (\textcolor[rgb]{1,0,0}{red}) or without       (\textcolor[rgb]{0,0,1}{blue}) Ad$^2$Attack.
		The top row is a clean image where the target is being correctly tracked. 
		The search patch is input into the direct downsampling module (DiD) and super-resolution upsampling module (SrU) in the bottom row, where the tracker is blinded by the attacked image.
	}
	\label{fig:fig1}
\end{figure}
Specifically, as seen in Fig. \ref{fig:fig1}, the original image is directly downsampled (DiD) to lose some pixel features, where the multiple of the down-sampling is determined by the size of the target.
Subsequently, the LR images are resampled using a carefully built super-resolution upsampling (SrU) network to generate adversarial examples against UAV tracking. 
Our loss function is designed based on the classification score map and the regression map of the feature map extracted from the search patch image.
To achieve a promising trade-off between efficiency and imperceptibility, the input of our Ad$^2$Attack is the search patch instead of the entire image.
To our knowledge, this is the first investigation on the image-resampling technology in adversarial attacks against UAV tracking. In summary, this work makes the following contributions: 

\begin{itemize}
	\item Inspired by image-resampling, this work proposes an efficient attack approach, which consists of two parts: directly downsampling the original image and upsampling using super-resolution to generate the attacked image.
	\item A residual spatial enhancement module is proposed, which can better extract the features and highlight the pixels for attack.
	\item The loss functions, \textit{i.e.}, score-reversal loss and box-drift loss, are carefully developed for confusing the target area with the background area, and leading to the candidate bounding box drift.
	\item Experiments are conducted on three well-known UAV tracking benchmarks, the results demonstrate that our method can considerably lower the precision of the most advanced Siamese network-based trackers. Real-world test also verifies the efficiency and effectiveness of our attack approach. 

\end{itemize}

\section{Related Works}
\subsection{CF-based UAV Tracking}
Object tracking approaches can be broadly divided into methods based on correlation filters (CF) \cite{danelljan2017eco},\cite{fu2020disruptor},\cite{zheng2021mutation} and methods based on convolutional neural network (CNN) \cite{cao2021hift},\cite{Li2019CVPR},\cite{he2018twofold}.
Due to their efficiency\cite{li2020autotrack}, \cite{fu2020disruptor}, CF-based trackers were considered as the promising choice for UAV tracking. In the beginning, MOSSE \cite{bolme2010visual} has drawn increasing attention to the CF-based tracker. Y. Li \textit{et al.} in \cite{li2020autotrack} proposed a more adaptive and robust AutoTrack with automatic spatio-temporal regularization. While the SOTA CF-based trackers generally perform well in terms of real-time performance on a single CPU, with regard to the precision and success rate, the CF-based tracker is inferior to the Siamese-based tracker. 
\subsection{Siamese-based UAV Tracking}
Because of the attractive tracking performance, and good balance between accuracy and efficiency, Siamese trackers\cite{fu2021onboard}, \cite{Li2019CVPR},  \cite{guo2020siamcar}, \cite{guo2021graph} have become the current trend in visual tracking tasks.
SiamFC\cite{bertinetto2016fully} was a pioneering work, which proposed an end-to-end tracking strategy based on the use of a fully-convolutional neutral network to learn similarity.
In addition, SiamRPN++\cite{Li2019CVPR} adds clipping and spatial-aware sampling strategies in the residual unit, which significantly improves the performance of the Siamese tracker.  SiamAPN~\cite{fu2021onboard} designed an anchor proposal network for adaptive anchor proposal. As a continuation of SiamAPN, SiamAPN++~\cite{Cao2021IROS} introduces a novel attentional aggregation network to handle semantic information variation. HiFT \cite{cao2021hift} proposes a brand-new lightweight hierarchical feature transformer for effective and efficient multi-level feature fusion.



In summary, the Siamese-based trackers provide impressive performance in terms of precision and efficiency, but they are vulnerable to small adversarial perturbations in the image. In the process of training these trackers, the impact of adversarial examples is not considered, which means that the image after the adversarial attack has fatal damage to the robustness of these trackers.
\begin{figure*}[!t]	
	\centering
	\includegraphics[width=0.97\linewidth]{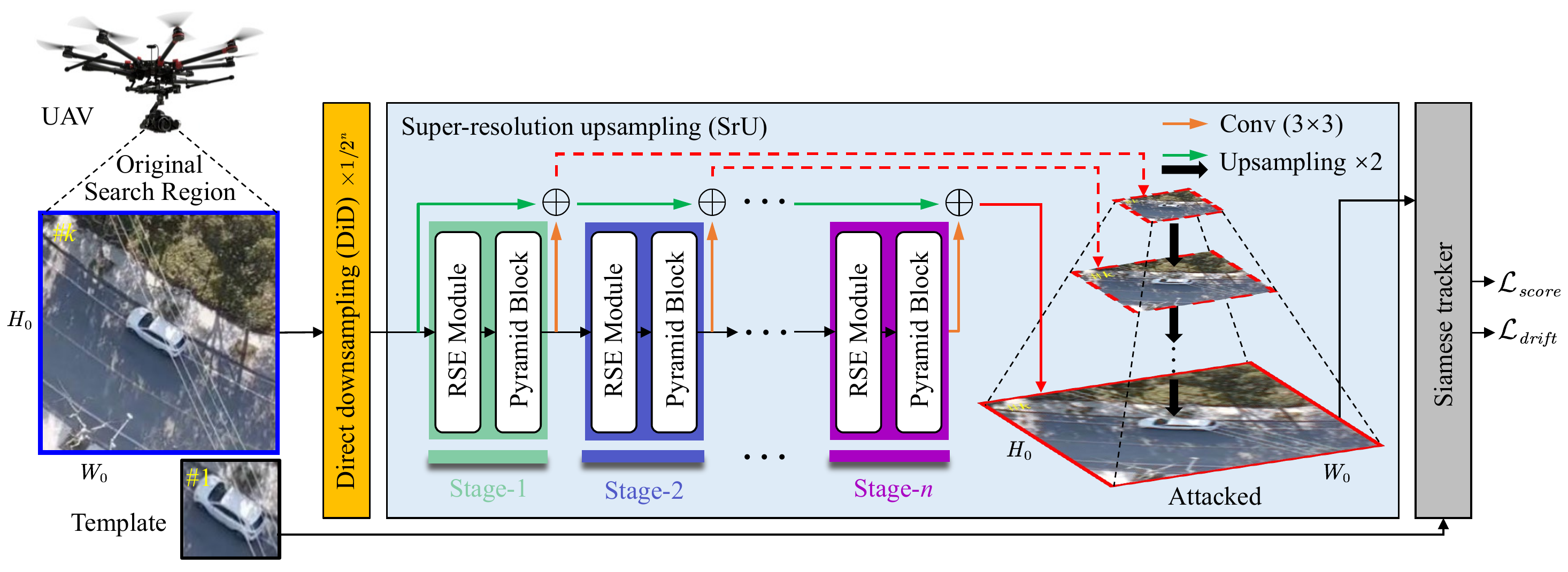}
	\setlength{\abovecaptionskip}{-1pt}

	\caption
	{
		Overview of our Ad$^2$Attack pipeline.
		The downsampling multiple is the same as the number of pyramid stage \textit{n}, which is determined adaptively by the search patch (\textcolor[rgb]{0,0,1}{blue}).
		The low-resolution image as the input of SrU module will be resampled to the attacked search patch image (\textcolor[rgb]{1,0,0}{red}). The adversarial example and the target template in the initial frame are input into the tracker, and train the SrU network through the loss function to achieve a balance between imperceptibility and attack effect.
	}
	\label{fig:main}
	
\end{figure*}


\subsection{Adversarial Attack}
Numerous works\cite{athalye2018synthesizing}, \cite{moosavi2017universal} have shown that CNN is highly vulnerable to adversarial attacks, which aims to deliberately add some imperceptible perturbation to the original images, causing the deep network model to give an incorrect output with high confidence. 
Currently, the adversarial attack methods can be mainly categorized as optimization-based approach and one-step approach. The former ones fool CNN through optimizing the loss function based on gradient to generate adversarial examples.
FGSM\cite{goodfellow2014explaining} develops a fast gradient sign method that perturbs original images in the direction of the gradient. 
C\&W\cite{carlini2017towards} realized three attacks by optimization methods under different norms.
However, due to the high cost of computing, these methods hardly fulfill real-time requirements.
The latter category is dramatically faster than the former category because one-step attacks\cite{yan2020cooling},\cite{xiao2018generating} are usually trained offline and the operation of generating perturbations does not need to be repeated.

Currently, there are some methods were proposed to attack visual object tracking. PAT\cite{wiyatno2019physical} and SPARK\cite{guo2020spark} attack the tracking task through iterative optimization on original frames. 
One-shot\cite{chen2020one} uses a dual attention mechanism to optimize the loss function, only slightly perturbing the target patch in the initial frame can blind the SOTA Siamese trackers.
CSA\cite{yan2020cooling} trains an effective and efficient perturbation generator with a carefully designed loss, which aims to simultaneously cool hot regions where the targets locate on the heatmaps and force the predicted bounding box to shrink.
\section{Proposed Method}
In this section, we introduce our Ad$^2$Attack framework against the Siamese tracker. The search patch will lose part of the pixel information through the direct downsampling module (DiD), and then is added imperceptible attack samples during the restoration through the super-resolution, which achieves the purpose of deceiving the Siamese tracker.

\subsection{Pyramid SrU module for Attack} 
We propose the SrU module based on the pyramid framework with two branches: \textit{1)} Features extraction and \textit{2)} Adversarial image reconstruction. After direct downsampling, the low-resolution (LR) image is entered into the SrU module. In the \textit{k}-th frame, our model progressively reconstructs the adversarial example on the pyramid level \textit{n}, where \textit{n}~is the downsampling scale factor and will be adaptively set according to the size of the target in the (\textit{k-}1)-th frame.
To achieve the effectiveness of reconstruction, the smaller the target is, the smaller the downsampling scale factor \textit{n} is.

\noindent\textbf{Feature extraction branch:} In the \textit{n}-th layer, feature extraction is composed of the same components: a Residual Space Enhancement (RSE) Module, and a pyramid block with d convolutional layers, and a transposed convolutional layer. Each transposed convolutional layer has two output connections: \textit{1)} a convolutional layer to generate the adversarial residual image (see Fig. \ref{fig:main}) and \textit{2)} directly fed into the more refined high-dimensional feature extraction branch of the (\textit{n+1})-th layer. By extracting features step by step from LR images instead of original images, and sharing feature representations in lower-level with that in high-level, the nonlinearity of the network can be increased to learn more sophisticated complex mappings, while significantly reducing calculations overhead.

\noindent\textbf{Adversarial image reconstruction branch:} In the \textit{n}-th layer of the pyramid, the LR image is upsampled \textit{×2} through bilinear interpolation, and the upsampled image is combined with the adversarial residual image generated from the feature extraction branch using an element-wise summation method to generate high-resolution (HR) images. Then the output HR image of the \textit{n}-th layer is used as the input of the (\textit{n+1})-th layer image reconstruction branch. It can be seen that the entire network is a cascaded CNN and each layer has the same structure. 

\subsection{Residual Spatial Enhancement Module}
\begin{figure}[!t]	
	\centering
	\includegraphics[width=0.9\linewidth]{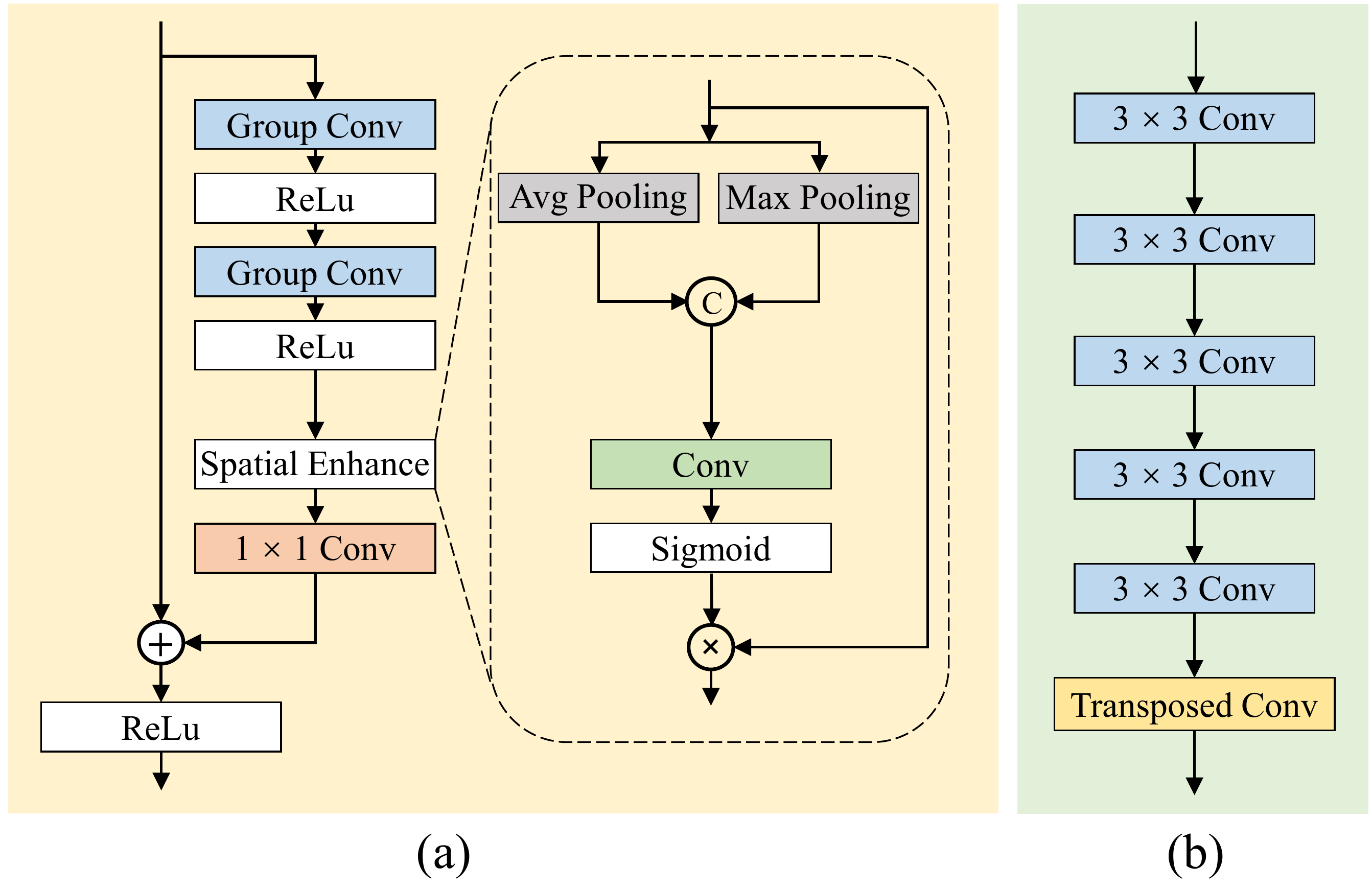}
	\setlength{\abovecaptionskip}{-0.08cm}
	\caption
	{
		Residual space enhancement module (a) and pyramid block (b). Attributing to the introduction of residual structure, spatial information is enhanced.
	}
	\label{fig:resffn}
\end{figure}

Most of the objects under the perspective of UAV have characteristics of small scale, various scale changes, and easy to be confused with the background. Consequently, this puts forward higher requirements for image spatial feature extraction. We propose the residual space enhancement(RSE) module as Fig. \ref{fig:resffn} to boost the performance of feature extraction from the target spatial location in the image. RSE is based on the residual structure. One of the branches uses group convolution to obtain complementary features of the channel and spatial latitude with lower coupling\cite{ioannou2017deep}. Meanwhile, filter relationships that don't have to be learned are no longer parameterized in the networks with group convolution. This can effectively reduce parameter quantity, and prevent the training process from overfitting. Inspired by CBAM\cite{woo2018cbam}, the spatial information of the feature maps is enhanced in this branch to make the residual images learned by the network more targeted for attack. The final 1\textit{×}1 convolution kernel and the subsequent activation function ReLu can greatly increase the nonlinear characteristics while keeping the feature maps size unchanged, and achieve cross-channel information interaction, which allows RSE module to get a more accurate and comprehensive image feature expression.

\subsection{Loss Function}
In the tracking process of the Siamese tracker, features of the template and search area are extracted through a specific backbone network. Based on these features, a classification score map $\mathbf{C}_{\rm i}^c \in \mathbb{R}^{H \times W \times D} $ and a regression map $\mathbf{R}_{\rm i}^c \in \mathbb{R}^{H \times W \times M} $ will be output. If the two maps are eventually disturbed by the adversarial examples, the tracking results will produce inaccurate results and lose the target in subsequent frames. Toward confusing these two maps, we design the score-reversal loss and the box-drift loss. 

\subsubsection{Score-reversal loss.} The purpose of the score-reversal loss function is to reduce the confidence score value of the target area in the search area, increase that of the background area, which means to reverse the scores in these two areas. Concretely, after the classification score map is activated with the shape $[H,W,2]$ by softmax function, the target probability $\mathbf{P}_{\rm t}^ c$ and background probability $\mathbf{P}_{\rm b}^ c$ of each pixel in the search area are generated as:
\begin{equation}\label{1}
    [ \mathbf{P}_{\rm b}^c, \mathbf{P}_{\rm t}^ c]={\rm softmax} (\mathbf{C}_{\rm i}^c) \quad.
\end{equation}

Based on the probability $\mathbf{P}_{\rm t}^c, \mathbf{P}_{\rm b}^c$ and the threshold $\epsilon$ set in advance, we can get the target area and background area of interest in the original image:
\begin{equation}\label{2}
    \begin{split}
        \dot{\mathbf{S}}_{\rm b |\rm i}^c&=\mathbf{S}_{\rm i}^C [\mathbf{P}_{\rm b}^c <- \epsilon] \quad, \\
        \dot{\mathbf{S}}_{\rm t |\rm i}^c&=\mathbf{S}_{\rm i}^C [\mathbf{P}_{\rm t}^c >  \epsilon] \quad. \\
    \end{split}
\end{equation}

We define the  score-reversal loss based on the difference between the confidence scores of the target and the background in search regions in both original and attacked images as follows :
\begin{equation}\label{3}
    \begin{split}
        \mathcal{L}_{score}=\phi \frac{1}{N} [(\mathbf{P}_{\rm t}^a[\dot{\mathbf{S}}_{\rm t|i}^c]&-\mathbf{P}_{\rm t}^c [\dot{\mathbf{S}}_{\rm t|i}^c])\\
         +(\mathbf{P}_{\rm b}^a[\dot{\mathbf{S}}_{\rm b|i}^c]&-\mathbf{P}_{\rm b}^c [\dot{\mathbf{S}}_{\rm b|i}^c])] 
    \quad.
    \end{split}
\end{equation}

The first term in this objective aims to simultaneously decrease the target probability and the second term aims to increase the background probability in classification score map to facilitate deviating the tracker.

\begin{figure}[!t]	
	\centering
	\includegraphics[width=1\linewidth]{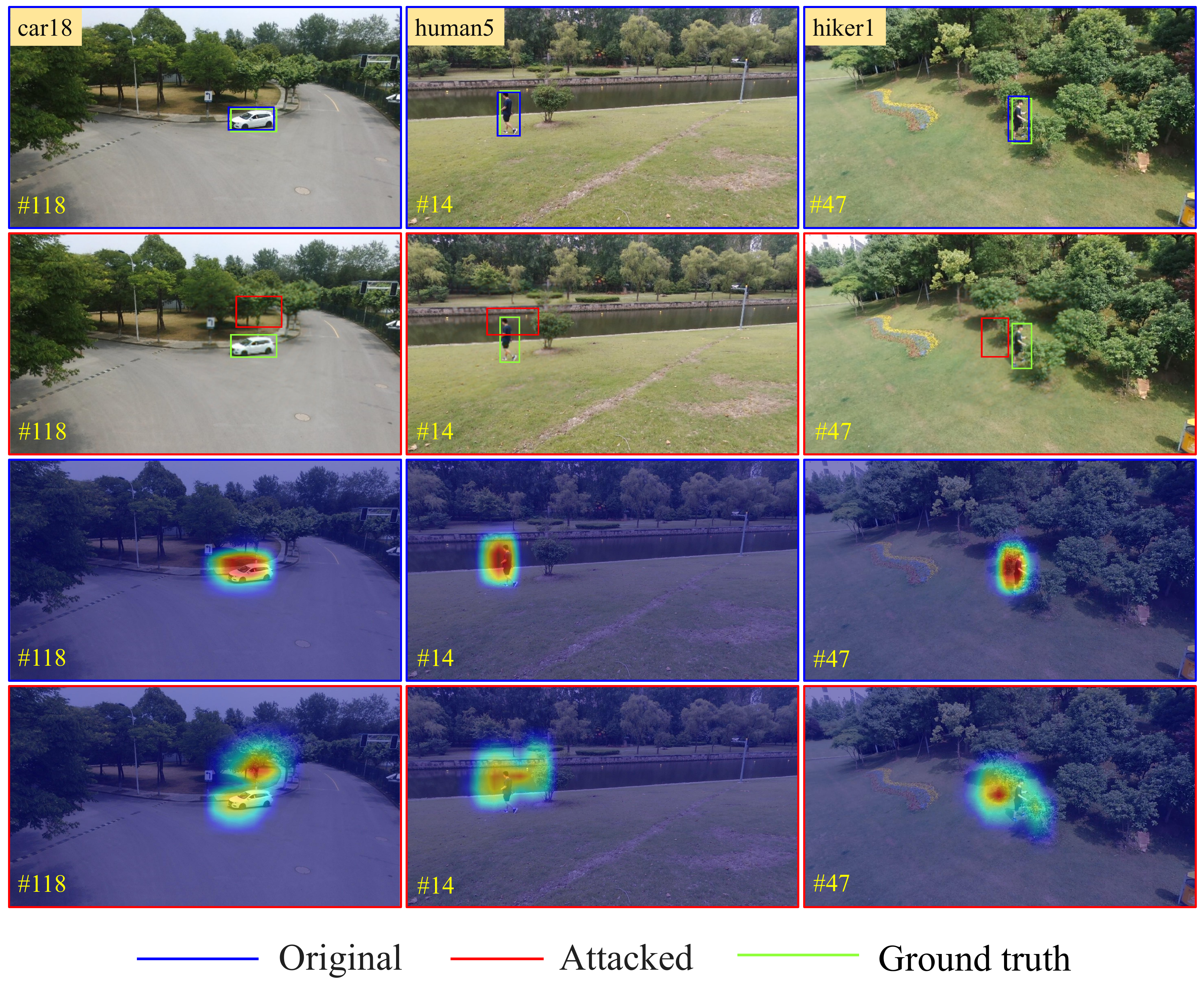}
	\setlength{\abovecaptionskip}{-0.5cm}
	\caption
	{
        Comparison of the output images and heatmaps with (\textcolor[rgb]{1,0,0}{red}) or without (\textcolor[rgb]{0,0,1}{blue}) Ad$^2$Attack. After attack, the high-response area that guides the target tracking will fade and deviate from the core part of the target.
	}
	\label{fig:4}
\end{figure}

\Remark Through back-propagating the logits output by the classification network of the selected tracker, \textit{i.e.}, HiFT, to the original image, the heatmap in Fig. \ref{fig:4} is produced, which reflecting the tracker's attention to the search region. As seen in Fig. \ref{fig:4}, this score-reversal loss makes the features of the target in the original image flattened, and the high-response area is enlarged. Therefore the precision of target tracking suffers a reduction.
\subsubsection{Box-drift loss. }The regression branch gets four tensors $\mathbf{R}_{\rm i}^a(x), \mathbf{R}_{\rm i}^a(y), \mathbf{R}_{\rm i}^a(w), \mathbf{R}_{\rm i}^a(h)$. The first two terms are used to adjust the center position of the bounding box, and the last two terms aim to adjust the size of the box. Our Box-drift loss is designed as follows:
\begin{equation}\label{4}
    \begin{split}
        \mathcal{L}_{drift}=\beta \frac{1}{N} \sum_{\dot{\mathbf{S}}_{\rm t|i}^c} {\rm max} (\mathbf{R}_{\rm i}^a(w)&+\mathbf{R}_{\rm i}^a(h),\tau_{\rm b})\\
        -\alpha \frac{1}{N} \sum_{\dot{\mathbf{S}}_{\rm t|i}^c} {\rm min} (\mathbf{R}_{\rm i}^a (x)^2 &+\mathbf{R}_{\rm i}^a (y)^2,\tau_{\rm c})
    \end{split}
    \quad,
\end{equation}
where the first term in the above formula is to shrink the box, and the second term is to force the center of the box to drift. $\beta$ and $\alpha$ are weight coefficients. The threshold values $\tau_{\rm b},\tau_{\rm c}$ set the degree of box size change and position offset in one frame. By optimizing the box-drift loss function, the error of the predicted bounding box will accumulate in terms of the scale and position, which makes the tracker fail.  

\Remark As can be seen from Fig. \ref{fig:4}, the box-drift loss makes the high response area deviate from the core position of the target, which reduces the success rate of the tracking.

\subsubsection{Overall loss functions. }In addition to the loss functions discussed above, a perceptibility loss $\mathcal{L}_2$ is used aiming to make the generated adversarial examples invisible to the naked eye. We express this loss as:
\begin{equation}\label{5}
    \mathcal{L}_{\rm 2}= \frac{\gamma}{N} \left\Vert \mathbf{S}^a-\mathbf{S}^c \right\Vert_2 \quad,
\end{equation}
where $\gamma$ represents the weight of this loss function. The complete objective is designed to train the SR network as: 
\begin{equation}\label{6}
    \mathcal{L} (T,z,S)=  \mathcal{L}_{score}+\mathcal{L}_{drift}+\mathcal{L}_{\rm 2} \quad.
\end{equation}
\Remark Through the adjustment of the weight value in the loss function, the attack can achieve a particular purpose. Specifically, increasing the weight value $\phi$ of the score-reversal loss can reduce the possibility of the target, so that the prediction box is enlarged and the precision rate is reduced.
\vspace{0.5cm}

\begin{figure*}[!t]	
	\raggedright
	\includegraphics[width=0.325\linewidth]{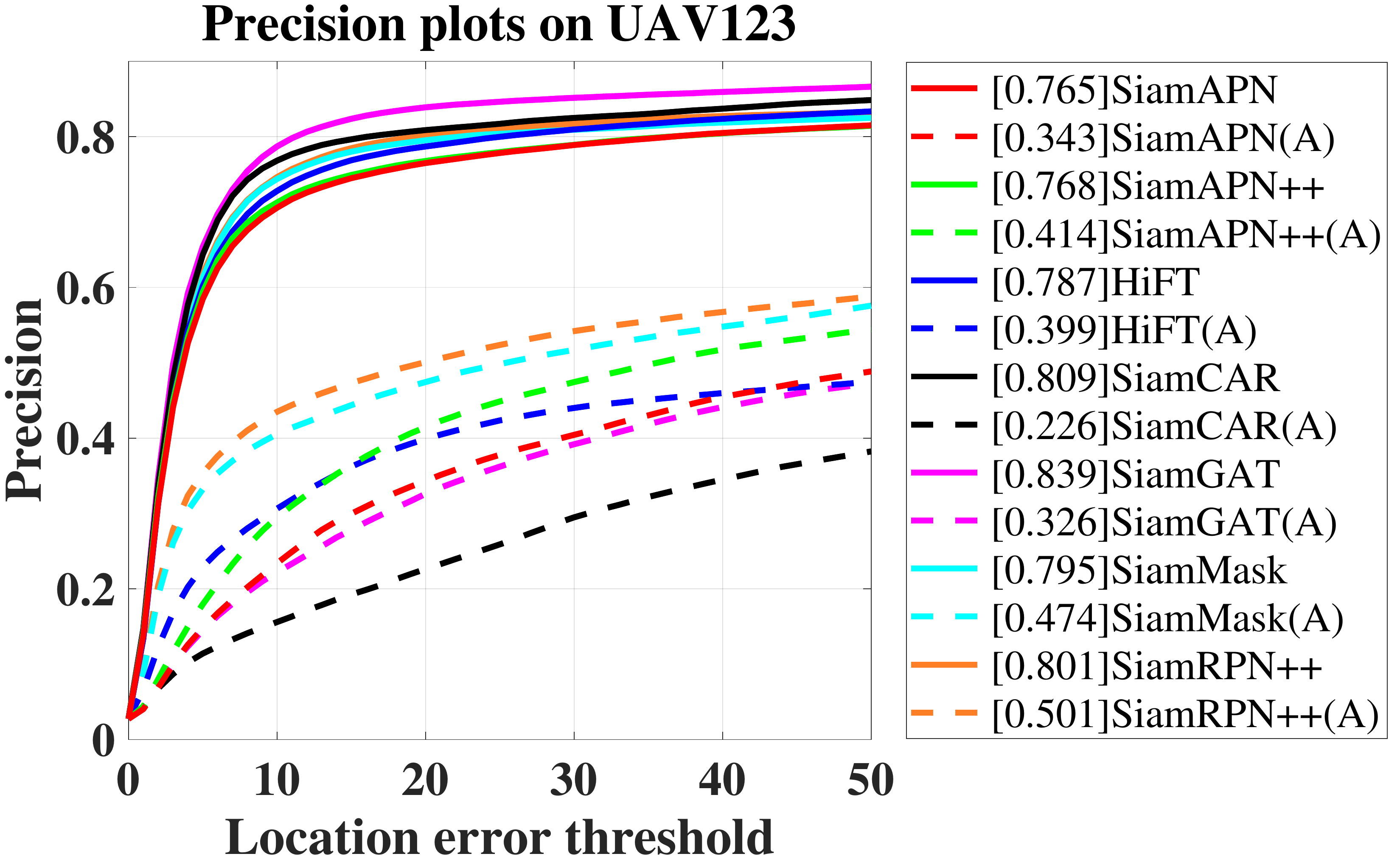}
	\includegraphics[width=0.325\linewidth]{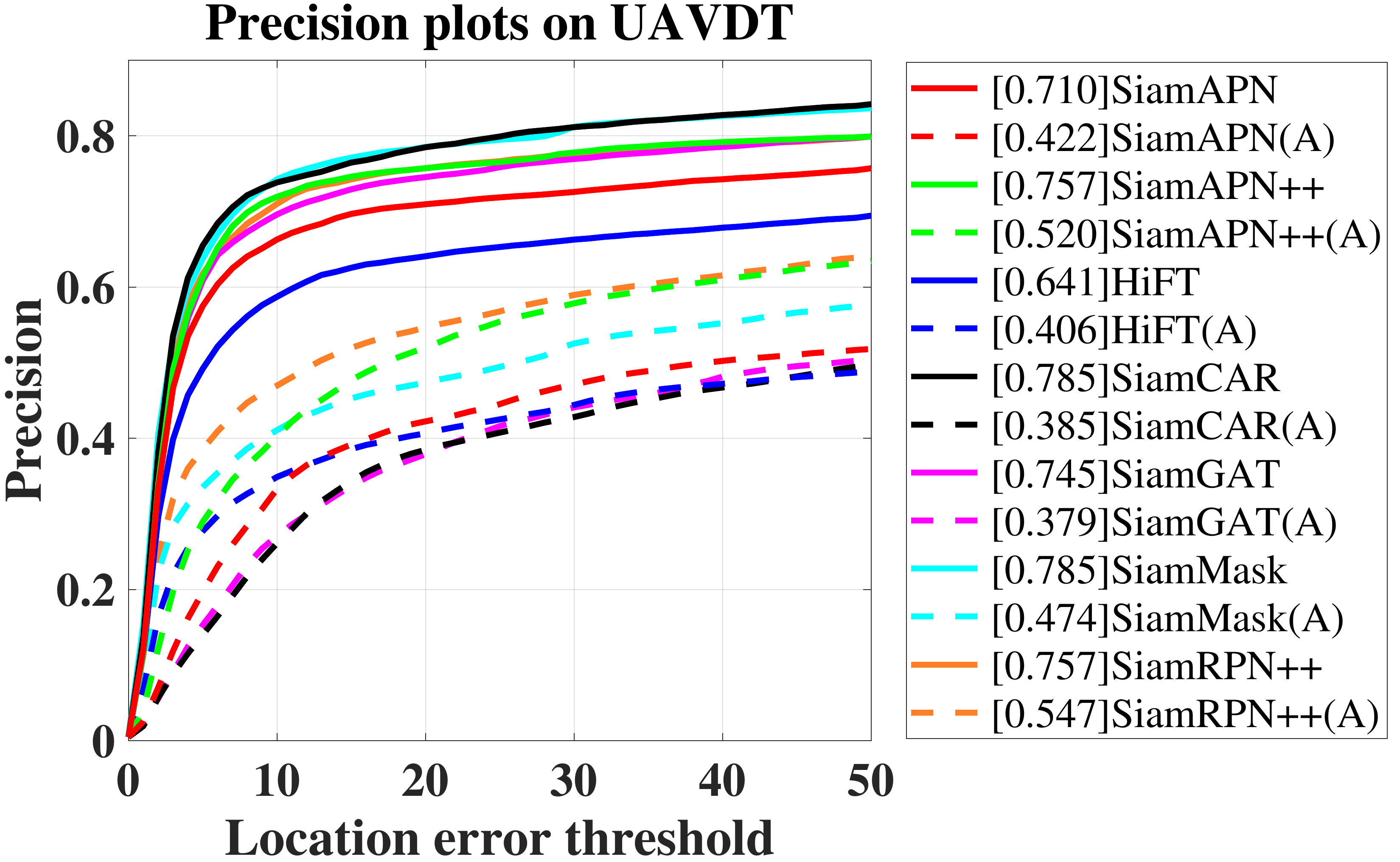}
	\includegraphics[width=0.325\linewidth]{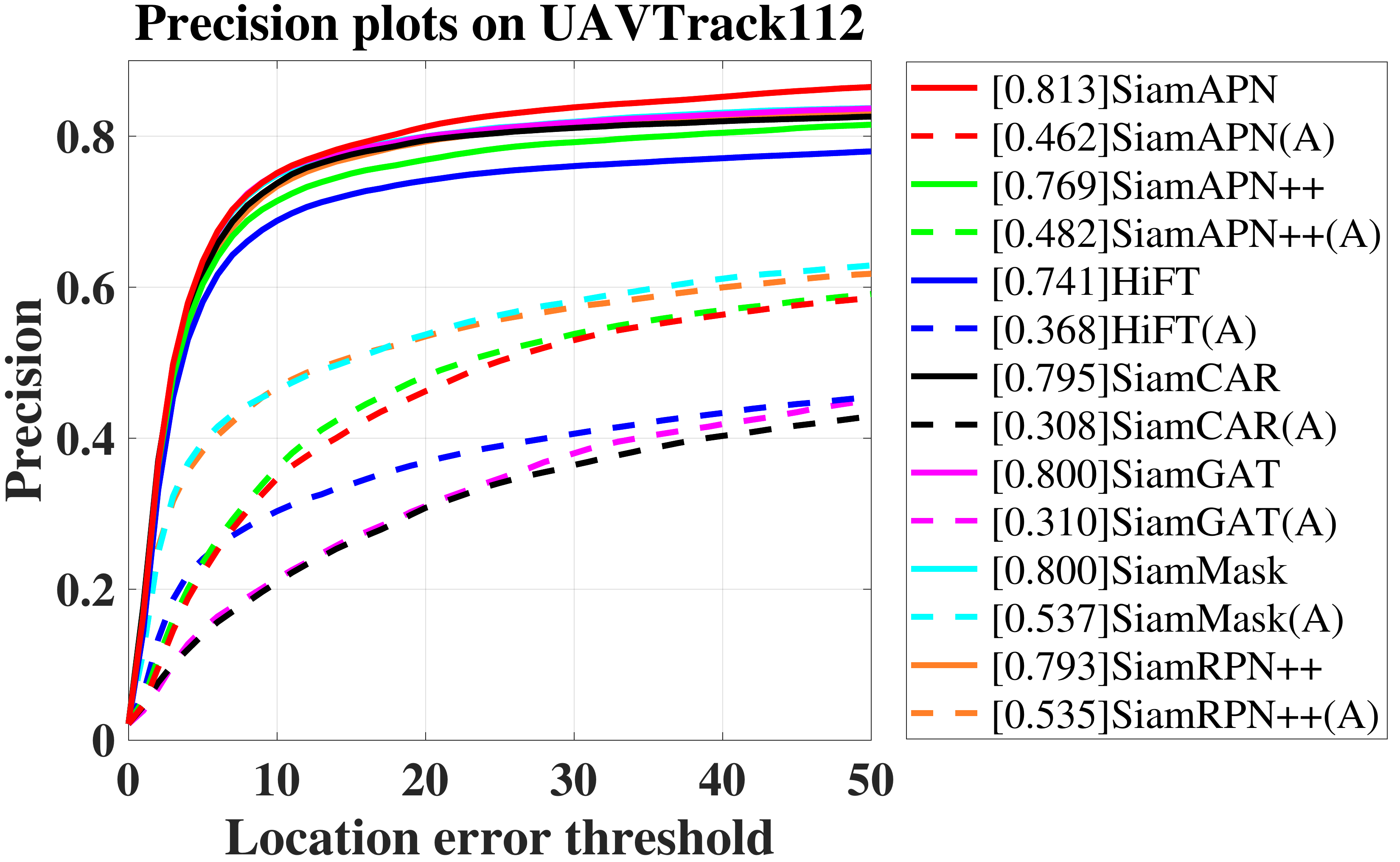}
    \label{fig:all-1}
    \vspace{-0.3cm}
\end{figure*}
\begin{figure*}[!t]	
	\includegraphics[width=0.325\linewidth]{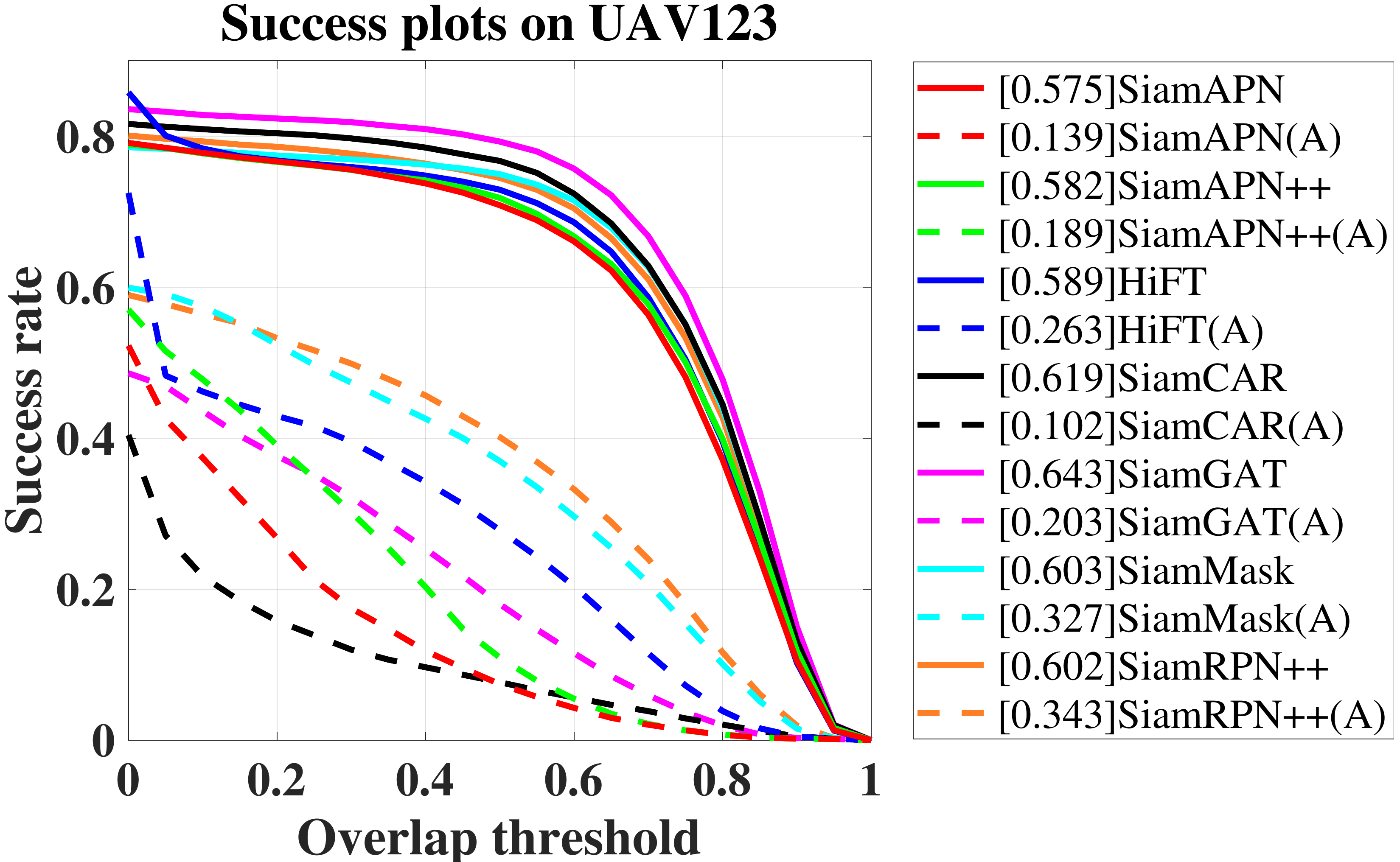}
	\includegraphics[width=0.325\linewidth]{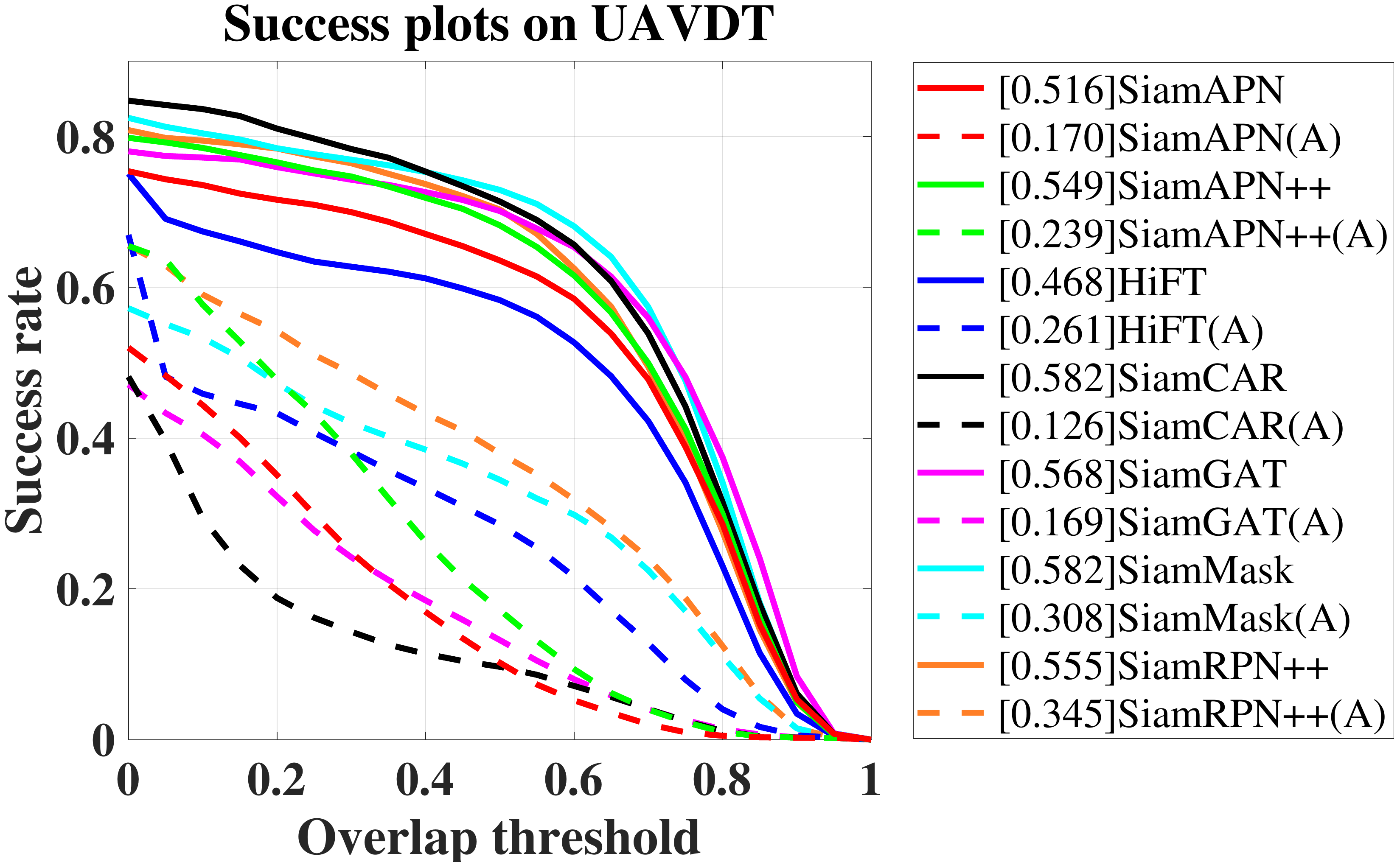}
	\includegraphics[width=0.325\linewidth]{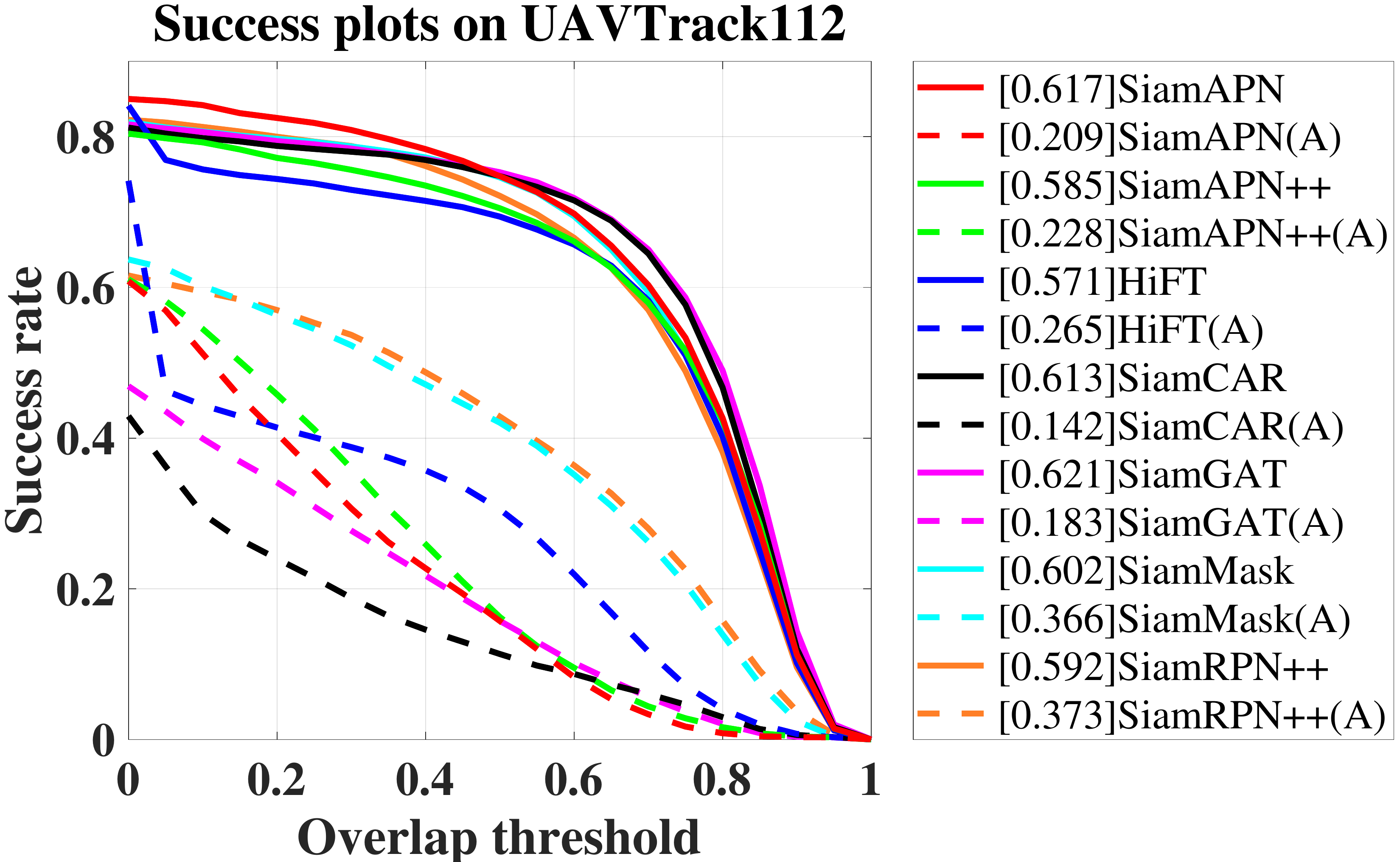}
	\setlength{\abovecaptionskip}{-2pt} 
	\caption
	{
		Overall performance of Siamese-based trackers with (dotted lines) or without (solid line) Ad$^2$Attack on UAV123, UAVDT, and UAVTrack112. Note that the precision and success rate of all trackers drop significantly after attack.
	}
	\label{fig:all}
	\vspace{-0.4cm}
\end{figure*}
\vspace{-1.5em} 
\subsection{Adaptive number of pyramid level ($N_{pl}$)}
Considering that too much downsampling will directly affect the image quality after super-resolution, causing the generated adversarial examples to be easily observed by human eyes, the scale factor of downsampling and the number of stages in SrU is designed to be adaptive to the target size and defined as a function below:
\begin{equation}\label{D1}
    \begin{split}
        N_{pl}&=\lfloor \sqrt{H_s \times Q_{i-1}} \rfloor \quad,\\
        Q_{i-1}&=\frac{h_s \times w_s}{H \times W} \quad,
    \end{split}
\end{equation}
where $\lfloor \star \rfloor$ is the largest integer not greater than value $\star$, $H_s$ refers to the size of the search region that enters the tracker network, here is 255. $Q_{i-1}$ represents the proportion of the search region in the entire image in the previous  (\textit{k-1})-th frame. This can better adapt to the characteristics of scale changes of the target from the perspective of UAVs, and achieve a balance between super-resolution and attack effects. At the same time, it can reduce the amount of calculation in the actual scene and improve real-time performance.


\section{Experiment}
In this section, comprehensive quantitative experiments are performed to investigate two aspects: \textit{1)} we validate the effectiveness of our Ad$^2$Attack against current SOTA trackers on three public benchmarks, 
\textit{2)} by conducting ablation studies on various contrast experiments, we investigate and evaluate the contribution of each component of the SrU module. The code and demo videos are available at: \url{https://github.com/vision4robotics/Ad2Attack}. 

\subsection{Implementation Details}
We implemented Ad$^2$Attack with PyTorch and performed our experiments on NVIDIA Telsa V100GPU with 32GB RAM. We train our network by uniformly cutting out the search area every ten frames from the GOT-10K\cite{huang2019got} benchmark. Initially, the search area image is down-sampled by $2^\textit{n}$ times, and after going through the network, the adversarial example with the original resolution can be obtained. SiamRPN++ is used as the target Siamese tracker for training our attack method. Adam optimizer is used to optimize the effect of super-resolution and attack, and its learning rate is set to $2\times10^{-4}$. To better confuse the tracker, the edge thresholds $\tau_b$ and $\tau_c$ in Eq. (\ref{4}) are set to -5 and 10. The weight $\alpha=1$, $\beta=10$. In order to make the attack imperceptible to human eyes, the weight $\gamma$ of L2 is set to 700. 

\Remark The weight of the loss function $\alpha$, $\beta$, $\gamma$ has been carefully adjusted, so that the values of the three loss functions are in the same order of magnitude, which can balance the effects of image restoration and attack.
\subsection{Evaluation Metrics} 
For all four data sets, we deploy their common settings and use one pass evaluation (OPE) that covers two metrics, namely precision and success rate. The former is based on the intersection over union (IoU) between the ground truth bounding box and the predicted bounding boxes of all frames, while the latter is based on the center location error (CLE) between the true bounding box and the prediction. For details, please refer to \cite{wu2013online}. We use the decline of these two metrics as the evaluation metric of the attack effect.
\begin{table}[!b]
    \scriptsize
    \centering
    \caption{
    Average attacking results of our attack method against six SOTA trackers on three benchmarks. The biggest drop results are highlighted by bold red font.
    }
    \begin{tabular}{c|ccc|ccc}
        \toprule
        \multicolumn{1}{c|}{\multirow{2}{*}{Trackers}}&
        \multicolumn{3}{c|}{Precision}&
        \multicolumn{3}{c}{Success Rate}\\
        \multicolumn{1}{c|}{} & \multicolumn{1}{c}{Org.} &\multicolumn{1}{c}{Att.} 
        & \multicolumn{1}{c|}{$\Delta$} & \multicolumn{1}{c}{Org.} &\multicolumn{1}{c}{Att.}
        & \multicolumn{1}{c}{$\Delta$} \\
        \midrule
        \multicolumn{1}{c|}{SiamAPN}&0.763&0.409&-46.37\%&0.569&0.173&-69.67\%\\
        \multicolumn{1}{c|}{SiamAPN++}&0.765&0.472&-38.27\%&0.572&0.219&-61.77\%\\
        \multicolumn{1}{c|}{HiFT}&0.723&0.391&-45.92\%&0.543&0.263&-51.54\%\\        
        \multicolumn{1}{c|}{SiamCAR}&0.796&0.306&\textcolor[rgb]{1,0,0}{\textbf{-61.63\%}}&0.605&0.123&\textcolor[rgb]{1,0,0}{\textbf{-79.60\%}}\\
        \multicolumn{1}{c|}{SiamGAT}&0.795&0.338&-57.42\%&0.611&0.185&-69.70\%\\
        \multicolumn{1}{c|}{SiamMasK}&0.793&0.495&-37.60\%&0.596&0.334&-43.98\%\\
        \multicolumn{1}{c|}{SiamRPN++}&0.784&0.527&-32.67\%&0.583&0.354& -39.33\%\\

        \bottomrule
    \end{tabular}
    \label{tab:over}%
\end{table}

\subsection{Overall Performance}
\begin{figure}[!t]
	\raggedright	
	\includegraphics[width=1\linewidth]{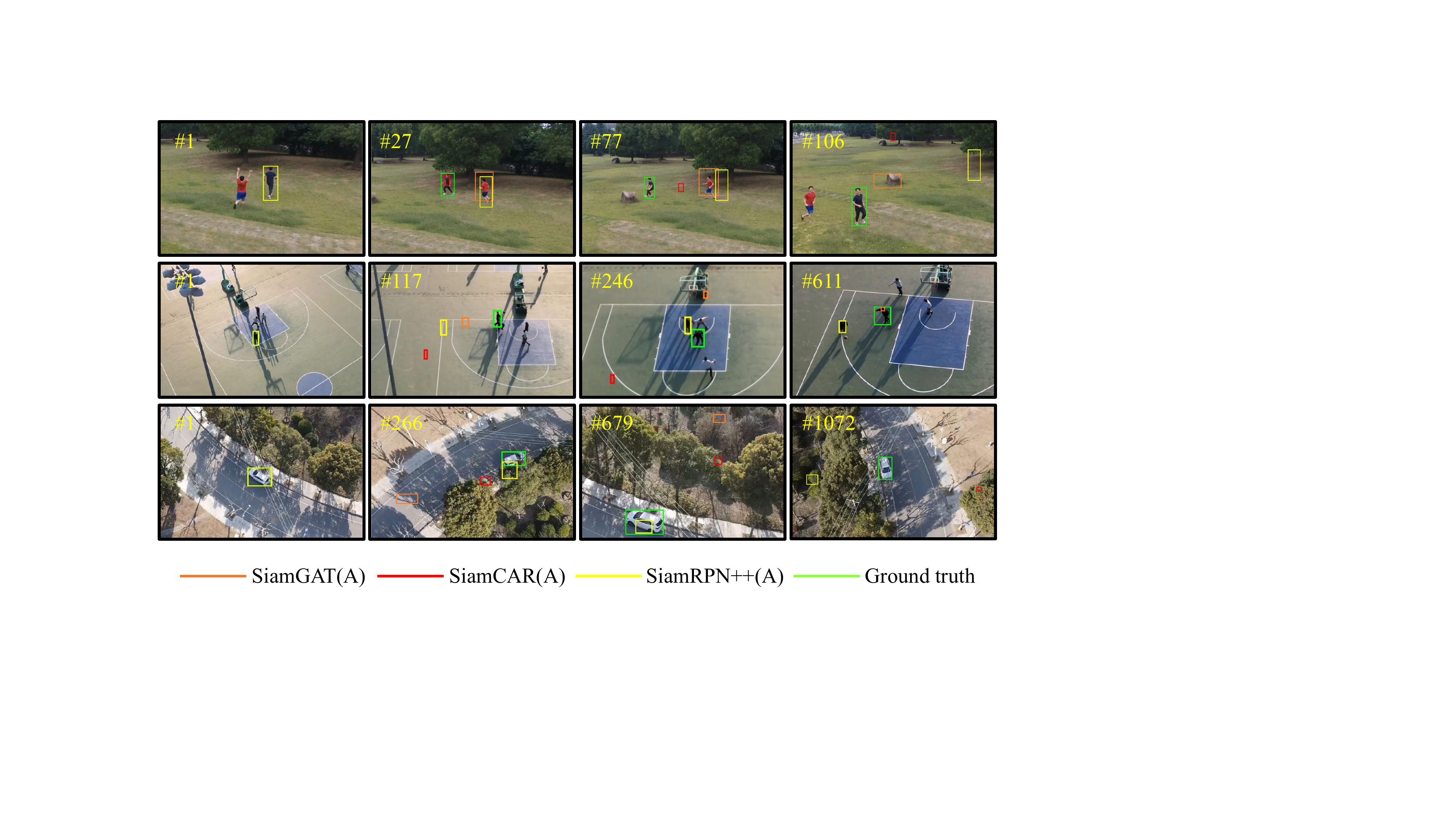}
	\setlength{\abovecaptionskip}{-0.4cm} 
	
	\caption
	{
		Some qualitative evaluation of trackers with attack. From top to down, the sequences are $human3$, $basketball$ $player1$, and $car4$ from UAVTrack112. With Ad$^2$Attack, the involved trackers fail to maintain robust tracking in these frames. 
	}
	\label{fig:qua}
\end{figure}

In this section we discuss the effectiveness of our attack methods on SiamRPN++\cite{Li2019CVPR}, SiamAPN\cite{fu2021onboard}, SiamMask\cite{wang2019fast}, SiamAPN++\cite{Cao2021IROS}, SiamGAT\cite{guo2021graph}, SiamCAR\cite{guo2020siamcar}, and HiFT\cite{cao2021hift}. Attack results on three public benchmarks are shown in Fig. \ref{fig:all}. 
It is observed that Ad$^2$Attack significantly reduces the precision and success rate of the seven trackers in three benchmarks. Specifically, on the UAVTrack112, Ad$^2$Attack reduces the precision and success rate of SiamRPN++ by \textbf{25.8\%} and \textbf{21.9\%}. In the three benchmarks, the attack effect on SiamCAR is the most obvious, and the precision and success rate dropped by more than \textbf{30\%}. It is because SiamCAR is a tracking strategy that determines the best target center point by observing the classification score map and the centrality score map, which is exactly the opposite of our loss function. The detailed tracking results after Ad$^2$Attack are shown in Fig. \ref{fig:qua} Average attacking results on these benchmarks are shown in TABLE \ref{tab:over}. It can be seen that the average performances of the trackers decrease enormously.

\begin{table}[!b]
    \centering
    \caption{	Qualitative comparisons between different combinations of modules in Ad$^2$Attack framework.
		Complete Ad$^2$Attack achieves the highest performance of attack.}
    \begin{tabular}{ccccc}
         \toprule
         \multicolumn{1}{c}{}&\multicolumn{1}{c}{Original}&\multicolumn{1}{c}{Down-up}&
         \multicolumn{1}{c}{W/o RSE}&
         \multicolumn{1}{c}{{Ad$^2$Attack}}\\
         \midrule
         \multicolumn{1}{c}{Precision}&\multicolumn{1}{c}{0.710} & 0.694 & 0.664 &\textcolor[rgb]{1,0,0}{ \textbf{0.422}}\\
         \multicolumn{1}{c}{Success}&\multicolumn{1}{c}{0.516} & 0.503 & 0.406 &\textcolor[rgb]{1,0,0}{ \textbf{0.170}}\\
         \bottomrule
    \end{tabular}
    \label{tab:1}%
\end{table}

\subsection{Ablation Study}
In this section, we analyze the impact of each module proposed in our Ad$^2$Attack framework. To this end, we use the SiamAPN tracker to perform an attack on UAV object tracking on the UAVDT\cite{du2018unmanned} benchmark. In TABLE \ref{tab:1}, we report our results with different combinations of modules in Ad$^2$Attack framework, \textit{i.e.}, tracking on the original images (Original), linear upsampled images after DiD (Down-up), Ad$^2$Attack without RSE module (W/o RSE), and complete Ad$^2$Attack. The results show that the merely linear upsampling of the image after DiD will lead to losing part of the detailed information, especially the edge information, which causes a decrease in the tracking results. In the absence of the RSE module, the picture only achieves global super-resolution and attack through several convolutional layers and transposed convolutional layers, which can further reduce the tracking results. And the complete Ad$^2$Attack uses the RSE module to emphasize the spatial information of the target in the search patch, so that it can attack the target tracking task in a targeted manner.

\section{Real-World Tests}
\label{sec:Real-WorldTests}
\begin{figure}[!t]	
	\centering
	\includegraphics[width=1\linewidth]{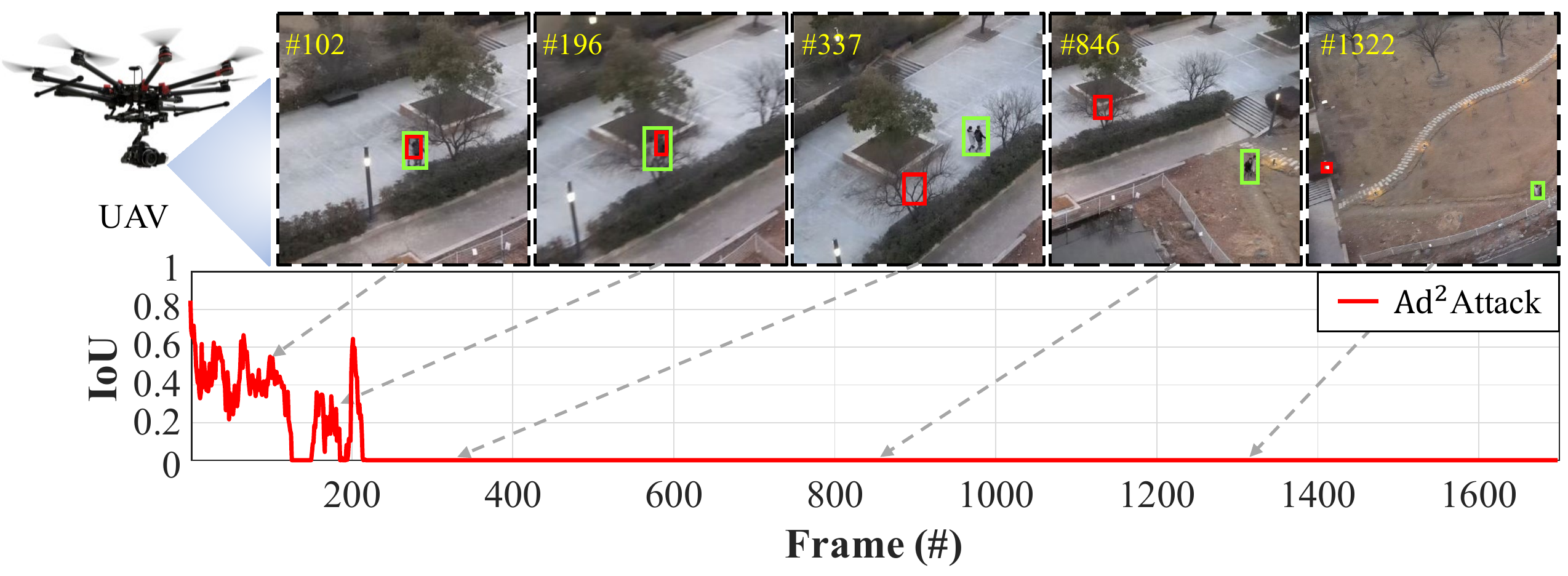}
	\includegraphics[width=1\linewidth]{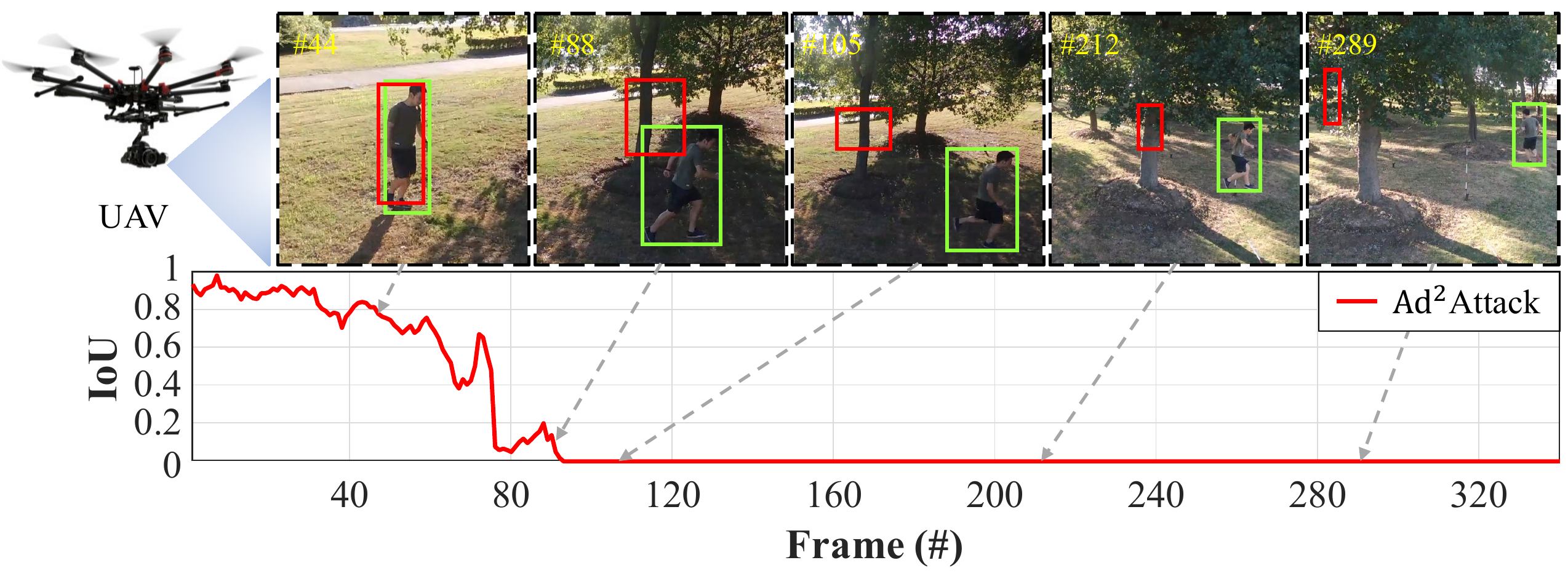}
	\includegraphics[width=1\linewidth]{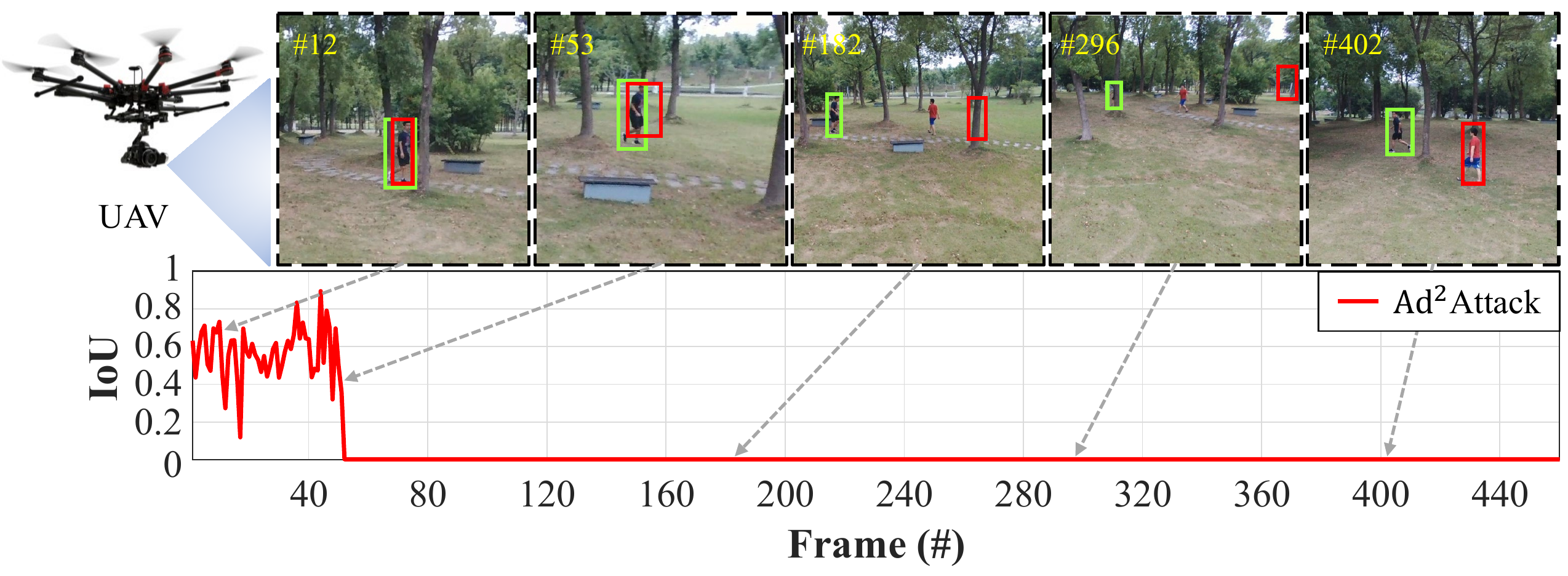}
	\setlength{\abovecaptionskip}{-0.4cm}
	\caption
	{
        Real-world test is carried out on a typical UAV platform. The (\textcolor[rgb]{1,0,0}{red}) and (\textcolor[rgb]{0,1,0}{green}) bounding boxes represent the tracking results after attack and the ground truth, respectively. The IoU curves are reported below the tracking snapshot. By attacking the search patch, the Siamese tracker produces poor performance.
	}
	\label{fig:real}
\end{figure}

Finally, a real-world test is processed on an NVIDIA Jetson AGX Xavier to verify the practicability and reliability of Ad$^2$Attack against UAV tracking.
We tested the efficiency and time cost of Ad$^2$Attack against SiamAPN. Fig. \ref{fig:real} reports the tracking results of the three tests in real-world. IoU curves between the predicted positions and the ground truth ones are illustrated, which can fully reflect the accuracy of tracking. The results in Fig. \ref{fig:real} demonstrate that Ad$^2$Attack causes severely tracking drift and fools the tracker to totally lost the object in a short period. Additionally, according to the real-world test on the UAV platform, our attack method can reach 72 frames per second without a great reduction of the tracking speed. It means that in terms of visual delay, our method can also be imperceptible to human eyes.

\section{Conclusions}
\label{sec:Conclusions}
In this work, for attracting more attention to the vulnerability and improvement of the robustness of UAV tracking model, an adversarial attack, \textit{i.e.}, Ad$^2$Attack, is proposed against UAV tracking. This method uses direct downsampling to lose pixel information, and then uses super-resolution technology to resample the image, which causes the tracking in subsequent frames to fail and make the attacked image imperceptible to human eyes. To better achieve the effect of super-resolution and attack, we propose the residual spatial enhancement (RSE) module to targeted express the feature of the image, and an elaborate loss function to drift the predicted bounding box. Experimental results prove that our method can successfully attack advanced Siamese trackers. To sum up, we believe that the proposed approach will considerably help reveal the drawbacks in UAV trackers and develop robust UAV tracking approaches.  


\section*{Acknowledgment}
This work is supported by the Natural Science Foundation of Shanghai (No. 20ZR1460100) and the National Natural Science Foundation of China (No. 62173249). 

\newpage
\bibliographystyle{IEEEtran}
\normalem
\bibliography{root}

\end{document}